\documentclass[pdflatex,sn-basic,Numbered,iicol]{sn-jnl}

\usepackage{xcolor}
\usepackage{amsmath,amssymb,amsfonts}
\usepackage{amsthm}
\usepackage{mathtools}
\usepackage{graphicx}
\usepackage{booktabs}
\usepackage{tabularx}
\usepackage{multirow}
\usepackage{enumitem}
\usepackage[strings]{underscore}
\usepackage{seqsplit}
\usepackage{xspace}

\newcommand{\SynAME}{\textsc{SynAM-E}\xspace}
\newcommand{\FlaDE}{\textsc{FlaDE}\xspace}

\hypersetup{hypertexnames=false,hidelinks}
\begin{document}

\title[SynAM-E: Event Cameras for Melt-Pool Monitoring]{Event Cameras for
Melt-Pool Monitoring in Additive Manufacturing: A Benchmark and a
Cross-Machine Transfer Analysis}

\author[1,2]{\fnm{Mohamad Yazan} \sur{Sadoun}}
\author[1,2]{\fnm{Sarah} \sur{Sharif}}
\author[3]{\fnm{Yingtao} \sur{Liu}}
\author[3]{\fnm{Zahed} \sur{Siddique}}
\author*[1,2]{\fnm{Yaser Mike} \sur{Banad}}\email{bana@ou.edu}

\affil[1]{\orgdiv{School of Electrical \& Computer Engineering},
\orgname{University of Oklahoma}, \orgaddress{\city{Norman}, \state{OK},
\country{USA}}}
\affil[2]{\orgdiv{INQUIRE Lab}, \orgname{University of Oklahoma},
\orgaddress{\city{Norman}, \state{OK}, \country{USA}}}
\affil[3]{\orgdiv{School of Aerospace \& Mechanical Engineering},
\orgname{University of Oklahoma}, \orgaddress{\city{Norman}, \state{OK},
\country{USA}}}

\abstract{\mathversion{normal}Melt-pool monitoring is central to qualifying metal additive manufacturing (AM), yet no public event-camera benchmark exists for this domain. Event cameras report per-pixel brightness changes with microsecond timing instead of reading full frames, giving the temporal resolution AM transients demand at a fraction of the data rate.

We present \SynAME{} (\emph{Syn}thetic \emph{AM} \emph{E}vents), the first public multi-source simulated event-camera benchmark for metal-AM melt-pool monitoring: 85 physics-calibrated event shards from 15 sources across 8 institutions, with public baselines and fixed cross-machine evaluation splits. On a single-machine case study, event-spatial monitoring matches dense-frame accuracy ($0.874$ versus $0.863$ macro-$F_1$), and the absolute intensity that events discard adds only $+0.006$ under fusion. On the NIST Additive Manufacturing Metrology Testbed (AMMT) build, a near-sensor event-rate counter recovers a raw-frame-confirmed $528.7$\,Hz intensity oscillation at ${\approx}380\times$ less sensor readout than the frame stream requires. A compact 93\,k-parameter spiking model runs at $15\times$ lower modeled inference energy for a $0.073$ macro-$F_1$ cost. Every cross-source task includes a built-in trust test against camera-identity shortcuts: process-type classification passes while material classification remains confounded by camera band, a corpus-structural limitation the release documents and the trust test exposes.}


\maketitle

\clearpage
\section{Introduction}
\label{sec:intro}

Metal additive manufacturing (AM) builds safety-critical aerospace
brackets, medical implants, and jet-engine parts. Its quality loop depends
on melt-pool monitoring that preserves process meaning across acquisition
setups and captures fast dynamics within practical data budgets. Two
challenges converge. First, in multi-lab data a monitor's score may reflect
the camera rather than the process. Second, porosity, lack-of-fusion,
keyhole instability, and spatter evolve at sub-millisecond timescales: a
keyhole cavity collapses in tens of
microseconds~\citep{zhao2017keyhole}, keyhole oscillation reaches
2.5--10~kHz~\citep{huang2022keyhole}, and spatter ejects at
1--20~m/s~\citep{leung2018insitu,ly2017metalvapor}. Frame monitoring can
reach 10~kHz, but a single build of \citet{lane2020ammt}'s Additive
Manufacturing Metrology Testbed (AMMT) Overhang part is already tens of
gigabytes at 8-bit grayscale.

Event cameras (dynamic vision sensor, DVS; dynamic and active-pixel
vision sensor, DAVIS) are an architectural alternative: each
pixel independently emits a timestamped $\pm 1$ event when its
log-photocurrent changes by a fixed contrast threshold, giving
microsecond timing, $>120$~dB dynamic range, and data sparsity
proportional to scene change, not scene
content~\citep{lichtsteiner2008silicon,brandli2014davispixel,gallego2022eventsurvey,chakravarthi2024survey}.
This architecture matches the AM regime: a bright emissive melt pool
against a dark substrate, sub-millisecond transients, and a latency
budget set by the laser-control closed loop. In practice the saving can
be large: recovering kHz-rate coherent dynamics from the event-rate
channel requires ${\approx}40$\,kB/s from a near-sensor counter, against
the $15.2$\,MB/s off-sensor readout any frame-derived channel must pay at
Nyquist (\S\ref{sec:results:deployment}).

\begin{figure*}[!ht]
\centering
\includegraphics[width=\textwidth]{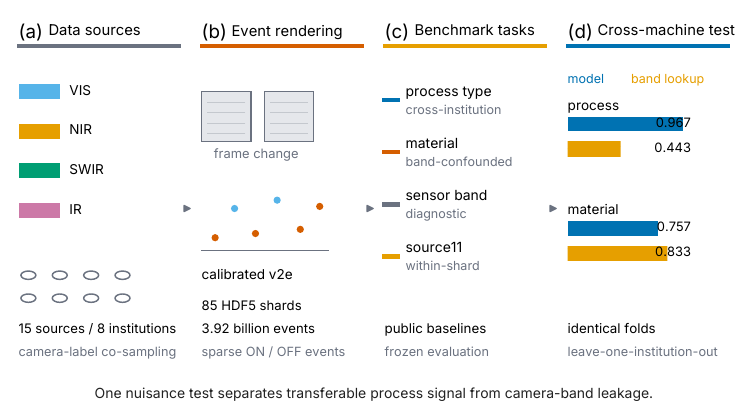}
\caption{\SynAME{} in one view. (a)~The problem: in multi-lab melt-pool data
each institution records with a single camera band, so labels travel with the
camera. (b)~What is released: an 85-shard, $3.92\times10^{9}$-event simulated
event-camera corpus (15 sources / 8 institutions, physics-calibrated v2e) with
public baselines and a frozen evaluation harness. (c)~What the test finds: on
identical source-keyed folds the trained model beats a camera-band lookup for
process type ($0.967$ against $0.443$, null $0.50$), while for the three source-keyed
material classes the lookup beats the trained SVM ($0.833$ against $0.757$;
strongest trained model $0.799$; null $0.333$): on those folds material is
not separable from camera-band identity (\S\ref{sec:results}).}
\label{fig:overview}
\end{figure*}

\SynAME{} provides testable evidence on both fronts
(Fig.~\ref{fig:overview}). On a single-machine case study, events match
dense-frame accuracy ($0.874$ vs $0.863$ macro-$F_1$); the event-rate channel
recovers a raw-frame-confirmed $528.7$\,Hz oscillation at
${\approx}380\times$ less sensor readout than the frame stream requires at
Nyquist; and a compact spiking model runs at $15\times$ lower modeled energy.
A built-in cross-site trust test, a no-training baseline that sees only
acquisition metadata on the same folds as the trained model, separates
process signals from camera-identity shortcuts;
leave-one-\emph{institution}-out splits keep each held-out laboratory
entirely out of training.

Prior event-based AM observation is confined to single
acquisition
setups~\citep{mascarenas2024demonstrating,springerjim2026memoryeff,bitharas2025oscillating},
leaving no public event corpus for cross-machine benchmarking. \SynAME{}
builds that basis on
\citet{jabbour2025meltpoolkinetics}'s multi-institution Melt-Pool-Kinetics
(MPK) video, following the event-vision practice that converted frame video
into event benchmarks before real captures were
plentiful~\citep{kim2021nimagenet,li2017cifar10dvs,gehrig2020vid2e}.

\paragraph{Contributions.}
\begin{itemize}[leftmargin=1.5em,itemsep=0.1em,topsep=0em]
  \item \textbf{C1: Public multi-source benchmark.} \SynAME{} is an
    85-shard, physics-calibrated v2e~\citep{hu2021v2e} re-rendering of the
    multi-lab Melt-Pool-Kinetics~\citep{jabbour2025meltpoolkinetics}
    release: 15 sources across 8 institutions, $3.92\times10^{9}$ events,
    per-shard provenance, a datasheet~\citep{gebru2021datasheets}, public
    baselines, and a reproducible evaluation harness. It is the first
    public, multi-source simulated event-camera benchmark for metal-AM
    melt-pool monitoring.
  \item \textbf{C2: Events match dense frames on one machine.} On a
    real-labelled single-machine thermal-anomaly task (source16),
    event-spatial monitoring reaches dense-frame accuracy across recordings
    ($0.874$ vs $0.863$ macro-$F_1$, 5 seeds) and clears an event-rate
    counter ($0.64$ fold-mean); the absolute intensity the events discard
    adds only $+0.006$ under early fusion (pre-release sweep;
    \S\ref{sec:results}).
  \item \textbf{C3: Deployment data rate and energy.} The event-rate channel
    recovers the corpus's fastest coherent dynamics, a raw-frame-confirmed
    $528.7$\,Hz oscillation on the NIST AMMT builds, at
    ${\approx}380\times$ less sensor readout than the frame stream
    requires. A compact 93\,k-parameter spiking model runs at
    ${\approx}15\times$ lower modeled energy for a $0.073$ macro-$F_1$
    cost on leave-one-source-out folds (\S\ref{sec:results}).
  \item \textbf{C4: Physics self-consistency.} A no-fit thermodynamic
    specialization of the \FlaDE{}~\citep{ding2025flade} event-rate model
    agrees with direct threshold counting across the 80-scene ideal grid
    (median ratio $1.07\times$; 80 of 80 within $2\times$; Online
    Resource~1, S6).
  \item \textbf{C5: Camera-band confound test.} In the released corpus, each
    contributing laboratory supplied one alloy through one camera band.
    A no-training baseline that sees only the camera band (a
    majority-label lookup per band, with no access to pixel data)
    classifies the three enrolled alloy classes \emph{better} than every
    trained model that sees the melt pool ($0.833$ against at best $0.799$
    macro-$F_1$ on identical leave-one-source-out folds; uniform chance
    $0.333$; permutation statistics in \S\ref{sec:results}).
\end{itemize}

\section{Related Work}
\label{sec:related}

\paragraph{Event sensors and simulators.}
\label{sec:related:sim}
Since the silicon-retina pixel of \citet{lichtsteiner2008silicon}, event
cameras have become commodity DAVIS~\citep{brandli2014davispixel,posch2014retinomorphic}
and Prophesee~\citep{perot2020gen4} families
(surveys: \citet{gallego2022eventsurvey,chakravarthi2024survey}). Because
paired frame (active pixel sensor, APS) and event data is costly, the field recycles video into events via
simulators: ESIM~\citep{rebecq2018esim,gehrig2020vid2e},
v2e~\citep{hu2021v2e}, DVS-Voltmeter~\citep{lin2022dvsvoltmeter},
V2CE~\citep{zhang2024v2ce}, ADV2E~\citep{jiang2024adv2e}, physics-first
PECS~\citep{han2024pecs}. \SynAME{} uses a calibrated v2e~1.7.0 pipeline whose
parameters are each a published DVS characterisation or documented calibration
choice (Online Resource~1, S5).

\paragraph{Sim-to-real for event cameras.}
\label{sec:related:sim2real}
{\sloppy
Prior studies quantify v2e-pipeline residuals in task-specific units:
segmentation-mIoU~\citep{jiang2024adv2e}, detection-mAP on
CARLA-DVS~\citep{tan2025carla}, point-cloud distance~\citep{han2024pecs},
downstream-task transfer~\citep{stoffregen2020reducing}. \SynAME{} tests
physics self-consistency (the no-fit-parameter event-rate envelope of
Online Resource~1, S6, derived in S8) and measures its residual directly against a real DVS
(\S\ref{sec:limitations}). DAVIS346 threshold/bias-physics limits are
characterised in
\citet{graca2023shining,graca2023optimalbias,delbruck2021feedback,graca2024autobiasing}.\par}

\paragraph{In-situ AM monitoring.}
\label{sec:related:am}
L-PBF defect monitoring is reviewed
in~\citep{grasso2017processdefects,grasso2021updatedreview,ng2024progress,cai2023insitu}; the
physics anchors for Online Resource~1, S8 (keyhole dynamics, 2.5--10~kHz
oscillations, spatter ballistics, melt-pool thermal scales, X-ray computed tomography (XCT) ground truth)
are \citep{zhao2017keyhole,hojjatzadeh2019pore,huang2022keyhole,leung2018insitu,ly2017metalvapor,bidare2018plume,heigel2020am,khairallah2016laser,praniewicz2020xct}.
Recent frame-based ML baselines report single-setup
accuracy~\citep{yang2024selfsupmelt,ren2023keyhole,transmatch2025}.
Within-machine in-situ monitoring is well covered: thermographic and
physics-guided classifiers detect porosity and process defects, supported by
melt-pool-image registration and in-situ/ex-situ data fusion, when training and test
share one acquisition setup~\citep{smoqi2022porosity,estalaki2022predicting,oster2023deepframework,kim2023meltpoolreg,kozjek2023datafusion}, but the cross-machine gap is well
documented~\citep{zhang2024domainshift,xie2024domainadaptation,beery2018terra}. Our cross-site trust test (\S\ref{sec:results}) quantifies
this gap for event-based monitoring.

\paragraph{Event precedents and scope.}
\label{sec:related:precedents}
No prior event-camera release targets metal-AM process monitoring with
public labels. The closest analogue is \FlaDE{}~\citep{ding2025flade}, a
real-DVS \emph{flame} benchmark whose event-rate equation we specialize
(Online Resource~1, S8) and whose BEC-SVM pipeline we reproduce
(Online Resource~1, S3). Adjacent demonstrations observe welding, a
liquid-gallium proxy, manual assembly, or concrete AM, without public
metal-AM
labels~\citep{mascarenas2024demonstrating,dold2025laserwelding,springerjim2026memoryeff,duarte2024edat24,mascarenas2026concrete};
\citet{bitharas2025oscillating} deploy a Prophesee EVK4 on an operating
L-PBF testbed but share data on request only, with no labels or benchmark;
\citet{starleaf2025hybrid} replay rendered test patterns on a 240\,Hz
display with no AM data; and neuromorphic-for-AM work deploys Loihi with
photodiodes~\citep{banerjee2025neuromorphic}, not an event camera.

\paragraph{Event-based ML baselines.}
\label{sec:related:evml}
The four families of Online Resource~1, S3 draw on time-surface
SVMs~\citep{lagorce2017hots,sironi2018hats}, event
CNNs~\citep{maqueda2018event,zhu2018evflownet,gehrig2019end,messikommer2020asynet},
event Transformers~\citep{sabater2022eventtransformer,gehrig2023rvt}, and
directly-trained SNNs~\citep{fang2023spikingjelly,neftci2019surrogate,zheng2021going,zhou2023spikformer,yao2024spikedriven};
established event benchmarks
(N-MNIST~\citep{orchard2015nmnist}, CIFAR10-DVS~\citep{li2017cifar10dvs},
DVS-Gesture~\citep{amir2017dvsgesture}, DSEC~\citep{gehrig2021dsec},
N-ImageNet~\citep{kim2021nimagenet}) and NeuroBench~\citep{neurobench2024}
(whose Pareto axes we follow) frame our protocol.
No prior work provides a public, labelled, multi-source event-camera corpus
for AM monitoring; \SynAME{} fills that gap.
\section{SynAM-E Dataset}
\label{sec:dataset}

\begin{figure*}[!t]
\centering
\includegraphics[width=\textwidth]{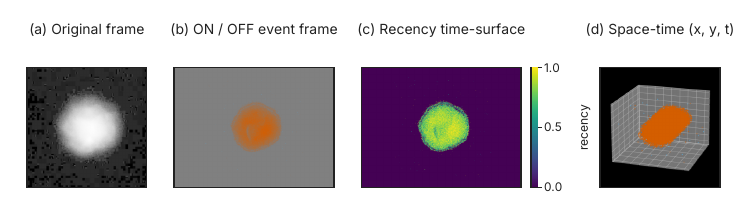}
\caption{One \SynAME{} melt-pool clip in four representations: (a) source
video frame; (b) ON/OFF event frame; (c) event recency time-surface; (d) the
raw space-time event cloud $(x,y,t)$. The event stream keeps the melt-pool
geometry and its sub-frame dynamics while discarding absolute intensity.}
\label{fig:representations}
\end{figure*}

{\sloppy
Every shard is generated from publicly released high-speed video by a calibrated v2e
pipeline~\citep{hu2021v2e} whose parameters tie to published DAVIS346
physics (Online Resource~1, S5); one parameter is deliberately pushed past its published bound
to stress-test throughput headroom (Online Resource~1, Table~S1). Construction is scripted.
The released shards use a single fixed-canvas configuration
(\texttt{DAVIS640} names the output canvas, not a separate sensor; the pixel
physics is calibrated to the DAVIS346). Figure~\ref{fig:representations}
shows one clip across the released representations.\par}

\begin{figure*}[!t]
\centering
\includegraphics[width=\textwidth]{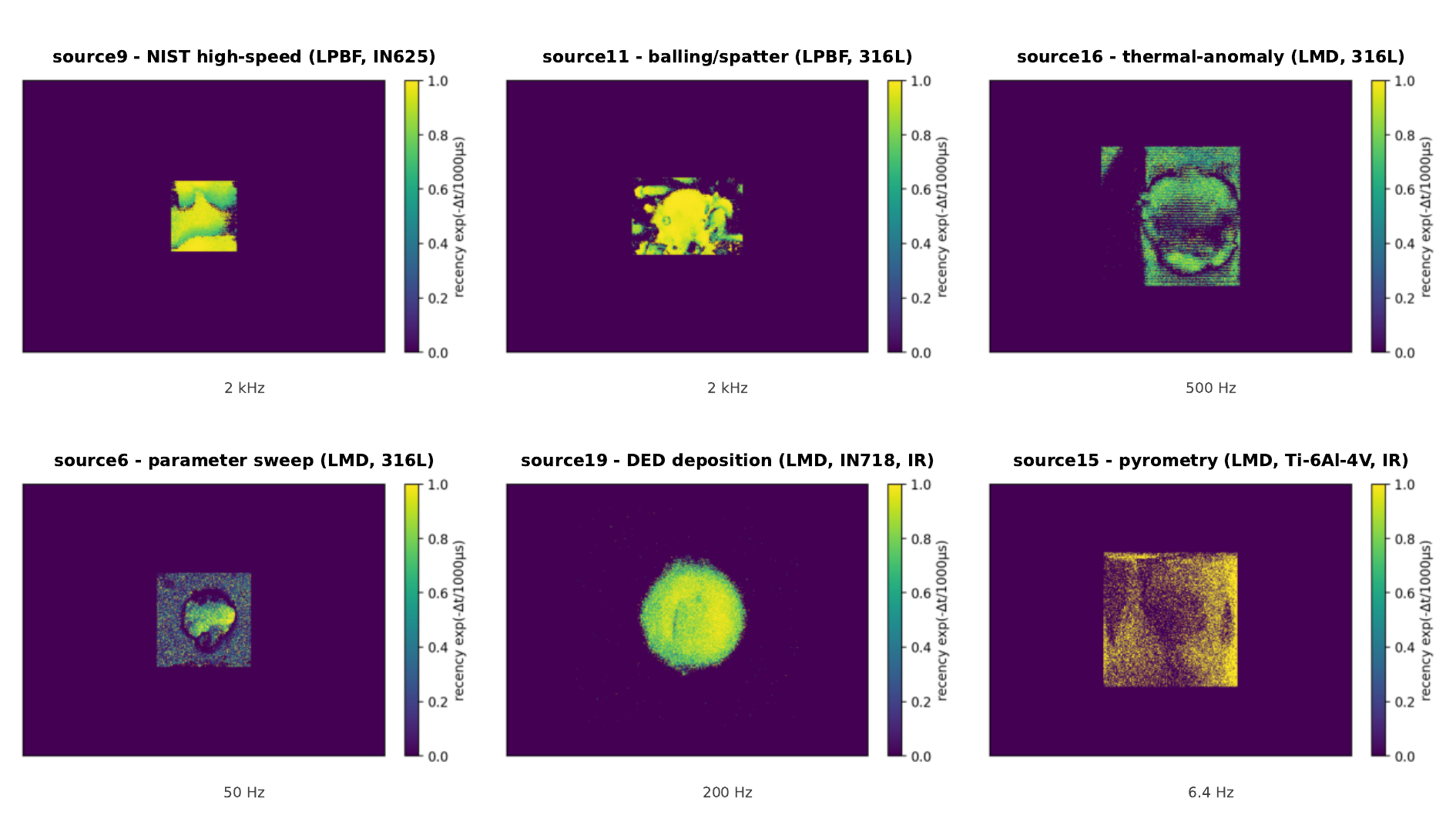}
\caption{Recency time-surfaces from six of the fifteen sources, one clip
each, spanning both processes (LPBF and laser metal deposition, LMD), acquisition rates from
6.4\,Hz (\texttt{source15} pyrometer view) to 2\,kHz (\texttt{source9},
\texttt{source11}), and camera bands
from visible through IR (\texttt{source15}, \texttt{source19}). All panels
share one colormap and scale; per-source contact sheets are included with the
release.}
\label{fig:gallery}
\end{figure*}

\paragraph{Source data and cross-institution cohort.}
The Melt-Pool-Kinetics corpus~\citep{jabbour2025meltpoolkinetics}
aggregates melt-pool video from independent AM institutions and instruments
(three sources captured on NIST Additive Manufacturing Metrology Testbed
hardware~\citep{lane2020ammt,lane2019ammt3d,yeung2018laser}); it provides frame video with structured
per-clip metadata but no event-camera data. The corpus comprises the 85
shards that render natively within the fixed DAVIS640 canvas; the release's
three oversized wire-arc AM (WAAM) records, from two labs, are excluded. The source
distribution is concentrated: 58 of the 85 shards come from one 50\,Hz source
(\texttt{source6}/UFPR), which is why every headline is a macro over folds,
never a clip-pooled mean. The enrolled tasks and institution-count rule are
in Table~\ref{tab:cohort}; the per-source composition is documented
for reproducibility (Online Resource~1, S1).
Figure~\ref{fig:gallery} samples the visual diversity this aggregation
produces.

\paragraph{Bandwidth calibration and provenance flags.}
\label{sec:dataset:stability}
v2e's first-order IIR photoreceptor is stable only for
$\varepsilon = 2\pi f_c \Delta t \leq 0.1$; we fix $\varepsilon = 0.05$ per shard by
construction (a $2\times$ margin) and set
$f_c^{\text{eff}}=\min(f_c,f_s/2)$. All 85 shards carry the resulting
\texttt{physical\_bandwidth\_compromised} and \texttt{nyquist\_capped}
flags. The two-tier policy and before/after analysis are in
Online Resource~1, S6.

\emph{Effective-bandwidth distribution and provenance.} Every shard records
$f_c^{\text{eff}}$ and provides its event stream, frame-index vector, metadata,
plan-file provenance, and checksum; the source distribution, kHz-capable
channels, and per-shard field inventory are in the released datasheet.

\paragraph{Border-artifact removal and event provenance.}
The released corpus removes the SuperSloMo zero-padding border artifact with a
2-px guard band, dropping $128{,}017{,}369$ events ($3.16\%$) and leaving
$3{,}923{,}454{,}180$; kept events are byte-identical to the unfiltered
parent (exact keep rule and verification gates: Online Resource~1, S1).

\paragraph{Tasks, splits, and dataset card.}
\label{sec:dataset:labels}
Labels, tasks, and headline folds are summarized in Table~\ref{tab:cohort};
Online Resource~1, S2 gives the exact split construction and Online Resource~1, S1 the
full datasheet-for-datasets~\citep{gebru2021datasheets}. The release, its per-shard
manifests, and the analysis artifacts are archived open-licensed on Zenodo.

\begin{table*}[!t]
\centering
\footnotesize
\caption{\SynAME{} cross-institution cohort (85 shards / 15 sources / 8
institutions). Headline classes are those carried by $\geq 3$ institutions;
counts trace to the released per-task leave-one-institution-out manifests.}
\label{tab:cohort}
\begin{tabularx}{\textwidth}{@{}l >{\raggedright\arraybackslash}X >{\raggedright\arraybackslash}X l@{}}
\toprule
Task & Headline classes ($\geq 3$ inst.) & $<3$-inst.\ / excluded & Eval \\
\midrule
process type   & LMD (multi-inst.) & LPBF (2 inst., underpowered); excl.\ WAAM (0 landed) & inst.\ hold-out \\
material       & 316L (4 inst.) & IN718 (SUPSI), IN625 (NIST), Ti-6Al-4V (MSU); all single-inst.; excl.\ Cp-Ti, AISI2205, AlMg3Cr (single-source classes) & inst.\ hold-out \\
sensor band    & VIS (4 inst.) & IR (2 inst.), NIR (2 inst.); underpowered; excl.\ MWIR, SWIR & inst.\ hold-out (diagnostic) \\
\texttt{source11} anomaly & balling / irregularity / normal / overheating (1/1 each) & --- & within-shard only \\
\bottomrule
\end{tabularx}
\end{table*}
\section{Experiments}
\label{sec:experiments}

\paragraph{Shared protocol.}
{\sloppy
Hypotheses and acceptance criteria are fixed before training, each experiment's
criterion is stated alongside its result, and every headline split is opened
once; post-hoc audit sweeps are reported separately (Online Resource~1, S7).
The source16 sufficiency and data-rate results
(\S\ref{sec:results:itworks},~\ref{sec:results:deployment}) use five seeds
$\{42,1,2,3,4\}$ with across-seed $95\%$ $t$-CIs; the cross-institution
BEC-SVM-AM and the modeled-energy comparison use three seeds
(mean\,$\pm$\,std); the physics sweep and the \FlaDE{} reproduction anchor
(Online Resource~1, S3, five seeds) use the percentile bootstrap
($2.5\%$/$97.5\%$, 10\,000 resamples). Full evaluation protocol, baseline
architectures and training, and runtime/hardware:
Online Resource~1, S2--S4 and~S6.\par}

\paragraph{Models.}
Four baseline families span the methods a practitioner would reach for
(defined in full in Online Resource~1, S3): BEC-SVM-AM, a classical event-feature
SVM (binned event-count features C1--C7, \FlaDE{}'s feature identifiers,
and a Nystroem-approximated radial-basis-function (RBF) kernel, a
scalable approximation that keeps the SVM tractable); ResNet18-event, an ImageNet-pretrained ResNet18 on
event voxel tensors (the event-spike-tensor (EST) representation); ViT-MPK, a fine-tuned ViT-B/16;
and ConvSNN, a directly-trained spiking ResNet (SpikingJelly LIF) in two
sizes: a deep SEW-ResNet18~\citep{fang2021sewresnet} and a compact
93\,k-parameter LIF-ConvSNN. The spiking baselines use leaky
integrate-and-fire (LIF) neurons over a $T{=}16$-step window, trained by
backpropagation-through-time with a surrogate gradient; each learned baseline
follows its source family's standard recipe (no \SynAME{}-specific tuning).

\paragraph{Ideal-model self-consistency (C4).}
\label{sec:experiments:e4}
The radiometric event-rate model (Online Resource~1, S8) is tested for internal
consistency over an $80$-setting ideal grid (temperature, cooling rate, DVS
threshold): the analytic contrast-variation prediction against direct
threshold counting on the same grey-body cooling field. Median ratio $1.07$;
$80$ of $80$ settings inside the $0.5$--$2.0$ band; the residual is a
one-sided $2.5$--$11.1\%$ discretization bias. No simulator enters this loop.

\paragraph{Event monitoring works (C2): the source16 protocol.}
\label{sec:experiments:itworks}
The single-machine sufficiency task is source16 (VUB MiCLAD, directed-energy-deposition (DED) 316L, NIR 500\,Hz) per-frame
thermal-anomaly detection, comparing three monitors \emph{at a matched
budget}: an event-rate scalar, an event-spatial CNN (ResNet18 on event
voxels), and the \emph{identical} ResNet18 on the dense frames the events
were rendered from, so the only variable is the representation. Splits are
contiguous time windows with a held-out gap; per-frame labels are
re-attached from the raw source and verified exact. The headline arm (ARM-B)
is leave-one-\emph{recording}-out on the shared $\{$Baseline,\,Edge$\}$
states on one machine (\S\ref{sec:results:itworks}).

\paragraph{Deployment: data-rate (C3).}
The data-rate sweep holds $n_{\text{train}}$ fixed per fold and varies
\emph{only} the deployed bytes/s (frame spatial/bit-depth budget; event
binning/threshold), so accuracy differences are attributable to the
representation's compressibility, not to sample count. It runs on the ARM-B
3-fold cross-recording task (same across-seed $95\%$ $t$-CIs); the axis is
event-vs-frame only.

\paragraph{Deployment: modeled energy (C3).}
\label{sec:experiments:e3}
We use the standard
synaptic-operation cost model~\citep{neurobench2024,horowitz2014energy}:
ANN energy $=$ multiply-accumulate (MAC) count $\times E_\text{MAC}$ for a single forward; spiking
energy sums, over the $T{=}16$ time steps the model executes, each layer's MAC
count scaled by its \emph{measured} mean spike rate at the per-accumulate
energy $E_\text{AC}$, with the
real-valued input layer charged at $E_\text{MAC}$ per the NeuroBench
first-layer convention; 45\,nm constants
$E_\text{MAC}\!=\!4.6$, $E_\text{AC}\!=\!0.9$\,pJ. The cross-model \emph{ratio}
is the reported quantity (decomposition: Online Resource~1, S5).

\paragraph{Cross-institution evaluation + the cross-site trust test (C5).}
The corpus-level generalization test is leave-one-\emph{institution}-out
(the 15 sources resolve to 8 institutions, \S\ref{sec:dataset}), with the
fold mean reported alongside the named worst held-out institution
(a pooled bootstrap would mask institution-specific variation). Within each
protocol every model family runs under identical folds and aggregation, so
comparisons are direct \emph{within} a table; the source-keyed cap-300 GPU
leaderboard (Online Resource~1, S5) is not comparable against the institution-keyed
rows of Table~\ref{tab:loso_svm}. The evaluation is gated by the confound
test of \S\ref{sec:results:audit}, fixed before reading model output
(Online Resource~1, S7).
\section{Results}
\label{sec:results}

\subsection{Single-machine sufficiency: events capture the monitoring signal}
\label{sec:results:itworks}

\paragraph{Cross-recording generalization (ARM-B).}
{\sloppy
Training on two recordings' shared $\{0,1\}$ states and testing the held-out
recording's $\{0,1\}$ on one machine, event-spatial monitoring is
\emph{at-least-as-good as dense frames in point estimate on every fold}
(Table~\ref{tab:armb}). Taking the three held-out recordings as the unit,
the per-fold differences give a mean $\Delta\!=\!{+}0.011$ with a $90\%$ CI
of $[{+}0.000,{+}0.022]$: no detected difference beyond the pre-declared
$\pm0.05$ macro-$F_1$ margin fixed before analysis (two-one-sided-tests (TOST) equivalence
logic;~\citet{schuirmann1987tost,lakens2017equivalence}). The gain over
the event-rate counter is spatial structure, not rate alone. The event representation discards absolute intensity entirely, yet the
monitoring signal survives: early fusion of the discarded frames adds only
$+0.006$ (fusion $0.878$ vs events-only $0.872$; frozen pre-release
ablation corpus, 5 seeds; Online Resource~1, S6), so absolute brightness
carries almost no information the temporal structure does not already
provide.\par}

\begin{table*}[!t]
\centering\small
\resizebox{\textwidth}{!}{%
\begin{tabular}{lrrccrl}
\toprule
held-out & majority & event-rate & event-spatial ($95\%$ $t$-CI) & frame ($95\%$ $t$-CI) & $\Delta$ & CIs \\
\midrule
587 & 0.432 & 0.627 & 0.811 [0.807, 0.814] & 0.797 [0.790, 0.804] & $+0.014$ & SEPARATED \\
588 & 0.418 & 0.615 & 0.911 [0.905, 0.917] & 0.895 [0.880, 0.910] & $+0.016$ & overlap \\
589 & 0.439 & 0.680 & 0.901 [0.900, 0.903] & 0.898 [0.896, 0.900] & $+0.004$ & abutting \\
\midrule
fold-mean & 0.43 & 0.64 & 0.874 & 0.863 & $+0.011$ & --- \\
\bottomrule
\end{tabular}}
\caption{ARM-B: leave-one-recording-out (released corpus) on the $\{$Baseline, Edge$\}$
states, one machine, 5 seeds, across-seed $95\%$ $t$-CIs ($\Delta$ from unrounded
means). Both spatial monitors clear the event-rate counter and majority
baseline; the event representation is sufficient: discarding absolute intensity costs nothing detectable.}
\label{tab:armb}
\end{table*}

\paragraph{Off-nominal screening across recordings (ARM-C).}
A third arm collapses the source16 thermal states to binary
nominal ($\{$Baseline, Edge$\}$) versus off-nominal and tests whether a
monitor trained on the other recordings flags deviant frames. Both spatial
monitors clear both baselines (majority and content-blind frame-index) on
every fold, with AUROC up to $0.9331$ (fold~590, frame monitor). One fold exposes a \emph{deviation-family boundary}: fold~589, trained without
overheat-family data, inverts for all three monitors, but adding fold~590
(strong-overheat) removes the inversion for both spatial monitors (frame
AUROC $0.3805\to0.7417$, event-spatial $0.3635\to0.7174$). A causal rolling
threshold holds the $1\%$ false-alarm budget on all nine fold/monitor
pairs, while a fixed training threshold realizes $21$--$38\%$ false alarms.
Controls (matched-$n_{\text{train}}$, event-rate counter sign flip) and
per-fold numbers: Online Resource~1, S6.

The within-sequence ARM-A results are in Online Resource~1, S6.

\subsection{Deployment characterization}
\label{sec:results:deployment}

Two separate axes: the deployed data-rate of the event stream
(event-vs-frame) and the modeled inference energy of the model family. All
reported event rates use the release's pre-readout evaluation plane (sensor
envelope and affected shards: Online Resource~1, S9).

\paragraph{(i) Data-rate.}
Events match the frame-best accuracy on all three folds (Online Resource~1, S6).
On 587 they also cost ${\sim}16\times$ less data; on 588/589 the event stream
needs ${\sim}1.5\times$ more data because the anomaly is coarse-spatial and
frames downsample gracefully while the melt pool is event-dense
(${\sim}5\times10^{6}$\,ev/s). The event edge is on the temporal axis: the
event-rate channel carries dynamics that no sub-kHz frame monitor can
represent, quantified in~(iii).

\begin{figure}[!htbp]
\centering
\includegraphics[width=0.85\linewidth]{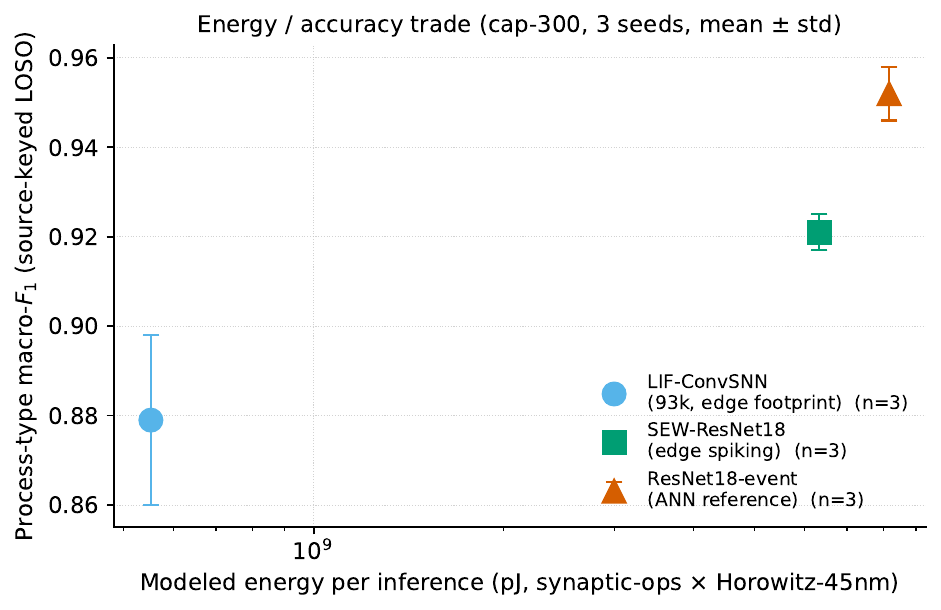}
\caption{Energy/accuracy trade on the cap-300 source-keyed leave-one-source-out (LOSO) leaderboard
(Online Resource~1, S5; 3 seeds, mean $\pm$ std). Model compactness is the efficiency lever: the 93\,k-parameter
LIF-ConvSNN spends ${\approx}15\times$ less modeled energy than the ANN
reference at a $0.073$ macro-$F_1$ cost; at matched architecture size the
spiking SEW-ResNet18 saves ${\approx}1.3\times$.}
\label{fig:pareto}
\end{figure}

\paragraph{(ii) Model energy and latency (model axis).}
\label{sec:results:pareto}
On the modeled NeuroBench / Horowitz-45nm synaptic-operation cost
(\S\ref{sec:experiments}), the efficiency lever is \emph{model footprint}
(Figure~\ref{fig:pareto}). The
compact LIF-ConvSNN (93\,k params) costs ${\approx}15\times$ less inference
energy than the ANN reference (ResNet18-event) at a $0.073$ macro-$F_1$ cost
($0.879$ vs $0.952$) and comparable measured latency ($1.1$ vs $0.9$\,ms per
sample, batch-1 fp32 on the reference RTX 5090). At matched architecture size, the spiking SEW-ResNet18 trades a $0.031$
macro-$F_1$ gap for a $1.3\times$ energy reduction: on the source-keyed
released 15-fold leaderboard
(Online Resource~1, S5, 3 seeds) the ANN reference reaches $0.952\pm0.006$,
SEW-ResNet18 $0.921\pm0.004$, and SEW spends ${\approx}1.3\times$ less energy at
${\approx}27\times$ the wall-clock latency ($24.0$\,ms per sample). Both source16 monitors are the same
ResNet18 with identical energy, so energy never enters the event-vs-frame
comparison; the accuracy is source-level (\S\ref{sec:results:source6};
cf.~\citet{banerjee2025neuromorphic}).

\begin{figure}[!htbp]
\centering
\includegraphics[width=\linewidth]{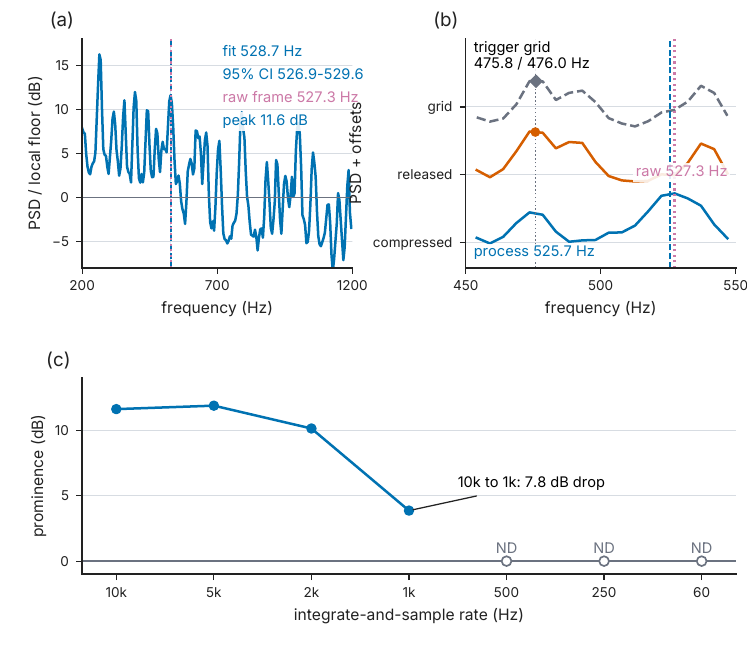}
\caption{The simulated event-rate channel recovers the fastest coherent dynamics
measured in the corpus, using the per-shard timing metadata the corpus
provides. (a)~\texttt{source8} (continuous clock): a $528.7$\,Hz line
(shaded $95\%$ CI), $11.6$\,dB over the local noise floor, coinciding with
the raw-frame difference line at $527.3$\,Hz (dotted).
(b)~\texttt{source10} (preserved trigger gaps): its $475.8$\,Hz peak matches
the manifest-only boundary train ($476.0$\,Hz, same FFT bin); deleting the
gaps moves the peak to $525.7$\,Hz, onto (a)'s raw-frame line.
(c)~Integrate-and-sampling \texttt{source8} at frame-camera rates (ND = not
detectable). All scalars are recomputed from the released shards.}
\label{fig:e5osc}
\end{figure}

\paragraph{(iii) Temporal fidelity: the corpus's fastest measured dynamics live in the event-rate channel.}
\label{sec:results:e5}
On \texttt{source8}, the simulated event-rate channel carries a coherent
intensity oscillation at $528.7$\,Hz ($95\%$ block-bootstrap CI
$526.9$--$529.6$\,Hz; Welch-averaged event-rate
spectra~\citep{welch1967psd}), $11.6$\,dB above a local colored-noise floor at
fixed spectral resolution (Figure~\ref{fig:e5osc}a). The released raw frames
(no event synthesis in the loop) put their frame-difference line at
$527.3$\,Hz, an agreement within spectral resolution, so the line is measured
source-video content and not a rendering artifact. It is a coherent line, not the maximum of an elevated
band: as resolution increases its $-3$\,dB width falls
$20.3 \to 8.5 \to 3.5$\,Hz while $Q$ rises $26 \to 62 \to 150$.

\paragraph{Cross-build consistency.}
Reading the other two builds requires the corpus's own timing metadata.
\texttt{source10} preserves $169$ native trigger gaps; on that clock its
dominant peak ($475.8$\,Hz) matches the manifest-only segment-boundary
impulse train ($476.0$\,Hz, same FFT bin; coherence $0.964$/$0.976$),
whereas on gap-free \texttt{source8} the same construction peaks at
$400.4$\,Hz, far from its event line. Deleting only the inserted gaps
recovers a stable $525.7$\,Hz line matching its own raw-frame $527.3$\,Hz
line. The same ${\sim}527$\,Hz line therefore
appears on both $10$\,kHz builds once each clock is handled correctly.

\paragraph{Spatial extent.}
Splitting the released raw frames into pixel-count-matched intensity
deciles, the line is present wherever the scene is bright enough to
resolve it ($7.5$\,dB top decile, $4.0$\,dB mid-brightness annulus),
with amplitude scaling with local brightness while fractional modulation
depth stays roughly constant ($0.019$--$0.041$). It falls below detection
in the near-dark region, a sensitivity limit rather than a spatial
boundary (injection and count-matching controls: Online Resource~1, S9).
The line is therefore a raw-frame-confirmed intensity oscillation
present across the lit field in proportion to local brightness, not
confined to the melt pool; assigning it a process mechanism requires gain
and laser-state metadata the corpus does not carry
(\S\ref{sec:limitations}). Per-region event-rate prominences:
Online Resource~1, S9.

A general methodological point: a periodogram computed across a preserved
trigger-gap clock manufactures a coherent line at the trigger-grid
frequency, so spectral claims on event datasets from triggered acquisition
require the per-segment gap structure \SynAME{} records in every shard
manifest.

The deployment consequence stands on \texttt{source8}, where the clock is
continuous. Integrating and sampling the same signal at conventional
monitoring rates collapses it (prominence $11.6 \to 3.9$\,dB, a $7.8$\,dB
drop on unrounded values, between
$10$\,kHz and $1$\,kHz; unmeasurable at ${\le}500$\,Hz,
Figure~\ref{fig:e5osc}c), so any sub-kHz frame monitor is blind to the
fastest coherent dynamics measured in this corpus. Recovering the line takes
only the event-rate channel: a $10$\,kHz counter at ${\sim}40$\,kB/s from a
sensor that never reads out a frame, against the $15.2$\,MB/s of off-sensor
readout any frame-derived channel must pay before the line is representable
(${\approx}380\times$; $120{\times}120$\,px $\times$ $8$\,bit at the
${\ge}1.06$\,kHz Nyquist rate). That $380\times$ is a sensor-readout ratio,
not a transported-bytes ratio (matched-bandwidth control: Online Resource~1, S9);
the detection margins are $8.6$\,dB for the counter against $0.88$--$4.5$\,dB
for every measured frame-derived channel. The counter presumes near-sensor
aggregation (an on-camera or FPGA counter binning the address-event
representation (AER) stream at
$100$\,\textmu s), bounded by the readout replay (\S\ref{sec:limitations}):
intact at the AER cap, lost at the 12~MEPS (million events per
second) host interface.

\subsection{Cross-site trust: separating process from camera}
\label{sec:results:audit}

Each contributing laboratory recorded with a single camera band,\footnote{Every
headline number in this section is computed on the released guard-band
corpus; the pre-release sweep numbers retained for context are labelled
where they appear.} so labels
and camera identity travel together. The cross-site trust test measures
whether a trained model's score reflects the process or merely the camera:
a content-blind baseline that maps each camera band to its majority label
(without examining any pixel data) is evaluated on identical folds
(Table~\ref{tab:audit}). A task the baseline reproduces is marked as
confounded; a task it cannot reproduce passes. Three apparent results are
identified as confounds before they enter headline tables
(Table~\ref{tab:audit}). The procedure adapts established confound-audit
practice~\citep{chyzhyk2022confound,torralba2011unbiased,spisak2022mlconfound,steinmann2024navigating}
and is released as runnable scripts with committed reference outputs.

\paragraph{Process-type classification passes the trust test.}
On leave-one-source-out folds, the camera-band baseline drops to
$0.443$ while the trained SVM scores $0.967$ (uniform chance $0.50$). The
baseline's $0.443$ also falls below its own content-blind nulls, so held-out
band membership predicts the wrong process type. The same
baseline keyed on resolution ($0.750$) and frame rate ($0.861$) also stays
below the SVM's $0.967$ on identical folds (Online Resource~1, S7).

\paragraph{The trust test catches a band confound in material classification.}
On identical leave-one-source-out folds, the camera-band baseline classifies the three
enrolled alloy classes at $0.833$ macro-$F_1$, above every trained model: the
BEC-SVM-AM reaches $0.757\pm0.020$, a ResNet18-event $0.770\pm0.016$, and a
linear reference $0.799\pm0.017$ (uniform chance $0.333$). Deeper models and
tuned hyperparameters do not close the gap (Online Resource~1, S7). The
confound is structural: each lab supplied one alloy through one camera, so
richer crossed band$\times$material coverage is needed before material
scores are interpretable. Permutation statistics: $p{=}0.0004$ treating the
14 material-labelled sources as exchangeable; $p{=}0.087$ under the
conservative institution-blocked null. Including the fourth enrolled class
(Ti-6Al-4V), which the band baseline structurally cannot reach, puts the
trained models above the baseline at every seed ($0.672$ event, $0.647$
frame, against $0.625$; Online Resource~1, S7).

\paragraph{Leave-one-institution-out evaluation.}
\label{sec:results:loso}
Two keying levels define the held-out folds: \emph{source-keying}
(leave-one-source-out, 15 folds) and \emph{institution-keying}
(leave-one-institution-out, 8 folds). Because the 15 sources map to only 8
institutions, source-keying can leave data from the same institution in
training, making it the appropriate level for the band-confound test above.
Institution-keying is the stricter cross-laboratory test
(Table~\ref{tab:audit}; weakest fold BAM $0.873$ seed-mean, worst seed
$0.857$). Under institution-keying, every fold holds out a single class, so
the fold-mean single-class $F_1$ ($0.966$) establishes survival under
institution holdout rather than discrimination; the two-class macro ($0.852$
at seed~0, $0.850$ three-seed) against a $0.50$ uniform chance indicates
discrimination in point estimate. LPBF sits at two institutions; a
three-institution design target follows (\S\ref{sec:limitations};
Online Resource~1, S9).

\paragraph{The confound is in the data, not the simulator.}
A natural question is whether v2e discards material-discriminative intensity,
disadvantaging the event-trained model. To test this, we trained a classifier
directly on the source video frames (raw $32\times32$ grayscale, the full
intensity range), on identical folds, with the identical Nystroem-RBF kernel.
The frame-trained classifier reaches $0.644 \pm 0.010$ on material, well
above the $0.402$ prior-weighted chance level but still below the
camera-band baseline's $0.833$. On process type the same frame pipeline
returns $0.743 \pm 0.020$ against the $0.443$ baseline, confirming
that the frame-side gate separates process type, not just material.
Under nested hyperparameter tuning, both event ($0.760 \pm 0.008$) and frame
($0.696 \pm 0.004$) models remain below the camera-band baseline
(tuning detail: Online Resource~1, S7).

\begin{table*}[!t]
\centering\footnotesize
\setlength{\tabcolsep}{3pt}
\begin{tabularx}{\textwidth}{@{}>{\raggedright\arraybackslash}p{0.17\textwidth} l >{\raggedright\arraybackslash}p{0.24\textwidth} l X@{}}
\toprule
task & keying & trained macro-$F_1$ & band lookup & null / verdict \\
\midrule
process (two-class) & source & SVM 0.967 & 0.443 & null 0.50; not reproduced by band alone \\
process (LMD $F_1$, fold-mean) & institution & 0.966\,$\pm$\,0.009 & 0.792 & single-class folds, null 1.0; survives institution holdout (weakest fold BAM 0.873) \\
process (two-class macro) & institution & 0.852 seed-0 (LMD 0.964 / LPBF 0.740); 0.850 three-seed & --- & null 0.50; point-estimate discrimination; $n{=}2$ interval spans chance (Online Resource~1, S9) \\
material & source & SVM 0.757\,$\pm$\,0.020; ResNet18-event 0.770\,$\pm$\,0.016 & 0.833 & null 0.333; \textbf{confounded} ($p{=}0.0004$ src-exch., $0.087$ inst.-blocked) \\
material (316L $F_1$) & institution & 0.405\,$\pm$\,0.047 & 0.500 & single-class folds ($p{=}0.54$); not cross-institution-testable \\
sensor band (VIS $F_1$) & institution & 0.452 & --- & self-referencing control (band $\equiv$ sensor); the low score reflects single-class institution folds \\
IR alloy slice & within-band & --- & --- & \textbf{confounded} by frame rate ($200$ vs $6.4$--$6.7$\,Hz, disjoint); no clean alloy test \\
\texttt{source11} 4-class & within-source & --- & --- & \textbf{confounded} by laser power; motivates the cross-institution framing \\
\bottomrule
\end{tabularx}
\caption{Cross-site trust grid: each task by keying, the trained score
against the content-blind band lookup on identical folds.
BEC-SVM-AM, 3 Nystroem seeds, mean\,$\pm$\,std; released guard-band corpus.
Per-keying statistics: Online Resource~1, S7. The grid separates
camera-independent results from confounded ones.}
\label{tab:audit}
\label{tab:loso_svm}
\end{table*}
\paragraph{Cross-machine feature robustness: a deployment guard.}
\label{sec:results:source6}
On specific held-out LMD labs the hand-crafted FlaDE features (BEC-SVM-AM)
\emph{transfer where deep models do
not}~\citep{geirhos2020shortcut,koh2021wilds}:
on \texttt{source6}/UFPR every deep model drops to
macro-$F_1\!\le\!0.20$ over 3 seeds (two of three families at exactly
$0.000$) while the features hold at $0.985\!\pm\!0.001$ source-keyed
($0.945$ institution-keyed), and the gap persists under a loosened
photoreceptor cap. This is the deployment guard (``when does a
within-machine monitor need hand-crafted features to transfer?'') and the source-level accuracy scope
behind the energy result (\S\ref{sec:results:pareto}).

\section{Limitations}
\label{sec:limitations}

\SynAME{}'s events are a calibrated re-rendering of real AM video, not a physical
event-camera capture. On a thermally-adjacent real-DVS proxy (FlaDE, real
DAVIS346; Online Resource~1, S9) the aggregate event rate is matched (median
ratio $0.94$) with a residual spatial-distribution mismatch
(Kolmogorov--Smirnov distance ${\approx}0.77$ per region); isolating sensor
transfer-validity requires paired real-DVS AM capture. The $528.7$\,Hz line
is recovered from the aggregate event-rate channel and confirmed in the raw
frames; the deployment energy is modeled (synaptic-ops $\times$ Horowitz-45nm).

\paragraph{Attribution of the $528.7$\,Hz line.} The line is raw-frame
confirmed and its amplitude scales with local brightness at a roughly constant
fractional depth, consistent with a melt-pool intensity oscillation. The
released shards carry no AMMT gain or exposure metadata, so the line is
reported without mechanism attribution; the near-dark null is a measured
sensitivity limit (injection control; Online Resource~1, S9).

\paragraph{Pre-readout event stream.} A measured readout replay
(Online Resource~1, S9) confirms the $528.7$\,Hz line survives the
${\sim}50$~MEPS AER cap ($1.76\%$ events dropped, prominence preserved) and
that the four source16 cross-recording shards behind the classification
results (ARM-B and ARM-C) lose at most $0.16\%$ of events under either cap,
so the reported macro-$F_1$ is unaffected; the released corpus does not model aggregate readout saturation,
but the replay bounds the consequence.

\paragraph{Scope of the sensor result.} The sufficiency finding (events match
dense-frame accuracy, Table~\ref{tab:armb}) holds on one machine (source16),
and the events are derived from the frames they are compared against; it
establishes representational sufficiency, not cross-machine or on-DVS transfer.

\paragraph{Institutional coverage.} The 15 sources resolve to 8 institutions,
with three classes each at a single institution. Process type is the one task
that survives institution-keyed evaluation; a class-stratified power analysis
(Online Resource~1, S9) locates the bottleneck at two LPBF-carrying
institutions and shows that a third closes the gap, which is the basis of the
data call in \S\ref{sec:conclusion}.

\section{Conclusion}
\label{sec:conclusion}

\SynAME{}, the first public multi-source simulated event-camera benchmark for
metal-AM melt-pool monitoring, establishes what event cameras deliver for this
domain. On a single-machine cross-recording case study, events match dense
frames in point estimate; the absolute intensity events discard adds only
$+0.006$ under fusion, so the temporal structure alone carries the monitoring
signal. The simulated event-rate channel preserves the corpus's fastest
raw-frame-confirmed dynamics (\S\ref{sec:results:deployment}) at
${\approx}380\times$ less sensor readout than the uncompressed frame stream
requires at Nyquist, dynamics sub-kHz frame monitoring cannot represent; the
trigger-gap analysis makes spectral claims on event datasets conditional on
the timing metadata each dataset must provide. On the source-keyed process
leaderboard, the compact 93\,k-parameter spiking model trades $0.073$
macro-$F_1$ for $15\times$ lower modeled inference energy. The benchmark's
built-in cross-site trust protocol tests each score against camera-identity
shortcuts: process-type classification passes while material classification
remains confounded by camera band, and leave-one-\emph{institution}-out
evaluation keeps each held-out institution out of training.

For a manufacturing team the results are concrete operating points.
Use the event-rate channel as an on-sensor counting mode where the
target is fast coherent dynamics at kB/s budgets (the densest build needs
near-sensor counting or 12~MEPS-plus interface headroom). Use dense frames
where absolute-intensity contrast carries the label. Use the compact spiking
model where the energy budget binds, at a stated $0.073$ source-keyed macro-$F_1$ cost.
Before trusting a pooled multi-site monitor, run the band lookup on identical
folds: a task the lookup reproduces needs crossed band$\times$material coverage
before any model result on it is interpretable. The single highest-value
addition to this corpus is one further LPBF-carrying institution, which
brings LPBF to the three-institution threshold this paper applies to its own
headline classes (\S\ref{sec:limitations}); labelled real-DVS recordings from
DAVIS346 or Prophesee hardware compose directly with the corpus released here
via a paired sim-vs-real protocol.
Together, the benchmark, the trust test, and the deployment measurements
establish, on simulated capture, that event cameras are a viable sensor
modality for metal-AM melt-pool monitoring: they match frame accuracy on
spatial tasks, uniquely access sub-millisecond dynamics at
kilobyte-per-second budgets, and expose cross-site shortcuts that pooled
frame data hides.

\bibliography{bibliography}

\section*{Statements and Declarations}

\bmhead{Funding}
The authors declare that no funds, grants, or other support were received
during the preparation of this manuscript.

\bmhead{Competing interests}
The authors have no relevant financial or non-financial interests to
disclose.

\bmhead{Ethics approval and consent to participate}
Not applicable. This work uses only publicly released video of inanimate
metal prints; no human participants or animals were involved.

\bmhead{Data availability}
The \SynAME{} dataset, per-shard provenance, and the analysis artifacts
behind every reported number are archived open-access on Zenodo
(CC~BY~4.0) at DOI \texttt{10.5281/zenodo.21359607}. The source video and labels derive from the
public Melt-Pool-Kinetics release~\citep{jabbour2025meltpoolkinetics}.

\bmhead{Supplementary information}
The supplementary document (Online Resource 1) contains the theory
derivations, methods and baseline details, split construction, the full
datasheet, and all supplementary tables and figures referenced in the text
as Online Resource~1, S1--S9.

\end{document}